\documentclass[conference]{IEEEtran}
\usepackage{graphicx}
\pdfoutput=1
\usepackage{float}
\usepackage{cite}
\usepackage{amsmath}
\usepackage{algorithm}
\usepackage{algorithmic}

\begin{document}

\title{Optimizing CNN Architectures for Advanced Thoracic Disease Classification}

\author{\IEEEauthorblockN{Tejas Mirthipati}
\IEEEauthorblockA{Georgia Institute of Technology, Atlanta, Georgia, USA\\
Email: tmirthipati@gatech.edu}
}

\maketitle

\begin{abstract}
Machine learning, particularly convolutional neural networks (CNNs), has shown promise in medical image analysis, especially for thoracic disease detection using chest X-ray images. In this study, we evaluate various CNN architectures, including binary classification, multi-label classification, and ResNet50 models, to address challenges like dataset imbalance, variations in image quality, and hidden biases. We introduce advanced preprocessing techniques such as principal component analysis (PCA) for image compression and propose a novel class-weighted loss function to mitigate imbalance issues. Our results highlight the potential of CNNs in medical imaging but emphasize that issues like unbalanced datasets and variations in image acquisition methods must be addressed for optimal model performance.
\end{abstract}

\IEEEpeerreviewmaketitle

\section{Introduction}
The integration of machine learning (ML) techniques in medical imaging has transformed traditional healthcare practices, enabling automated analysis of complex medical data. Among various ML techniques, convolutional neural networks (CNNs) have become a cornerstone for tasks such as image classification, segmentation, and object detection in medical imaging \cite{he2016deep}, \cite{rajpurkar2017chexnet}. CNNs are particularly effective in processing pixel data, making them ideal for analyzing chest X-ray images, a crucial diagnostic tool for detecting thoracic diseases such as pneumonia, tuberculosis, and cardiomegaly.

Despite their effectiveness, CNNs face significant challenges in the medical imaging domain, including dataset imbalance, noisy labels, variations in image quality, and complex multi-label classification tasks \cite{sethi2018chestx}, \cite{cichy2016deep}. In datasets like ChestX-ray14, these challenges are exacerbated by the prevalence of diseases such as "No Finding," which dominate the dataset and can lead to models that are biased towards detecting only the most frequent labels \cite{wang2017chestxray14}. Moreover, variations in imaging conditions—such as differences in X-ray positioning (front-to-back vs. back-to-front)—can introduce noise into the model training process, further complicating the task.

This paper presents a detailed study of CNN-based models for chest X-ray image classification. We investigate various architectures, including binary and multi-label classifiers, to classify a range of thoracic diseases. Additionally, we propose advanced preprocessing techniques, such as image resizing and PCA compression, to optimize model efficiency and reduce computational costs. To address class imbalance, we introduce a novel class-weighted loss function. We also explore the impact of additional metadata, such as patient demographics, on model performance, which has been an area of increasing interest in the context of personalized medicine \cite{gao2018deep}. This work aims to improve the accuracy, robustness, and generalizability of CNN models for medical image classification tasks.

\section{Related Work}
The application of deep learning techniques, particularly CNNs, to medical imaging has been extensively explored in recent years. One of the pioneering studies in this domain was by Rajpurkar et al. \cite{rajpurkar2017chexnet}, where a deep CNN model was developed to diagnose pneumonia from chest X-rays, outperforming radiologists in some cases. In a similar vein, \cite{sethi2018chestx} addressed multi-label classification problems in medical images, specifically in classifying chest X-rays into multiple disease categories. However, one of the most persistent challenges in this area is the inherent class imbalance in the datasets, where rare diseases are underrepresented, leading to biased predictions \cite{johnson2019survey}.

Recent work by \cite{luo2017multi} proposed an advanced multi-label classification framework designed specifically for medical image datasets, allowing for the simultaneous prediction of multiple diseases in a single X-ray image. This framework addressed the problem of missing data by employing advanced imputation techniques. Additionally, data augmentation and synthetic data generation have been explored to balance the dataset and improve model robustness \cite{shorten2019survey}. While these methods have been shown to enhance performance in some cases, they still face limitations, particularly in high-dimensional medical image datasets like ChestX-ray14.

In this study, we build upon these existing works by integrating class-weighted loss functions and leveraging PCA-based image compression to address dataset imbalance and reduce computational complexity. We also propose the inclusion of patient demographic information and additional metadata to improve the model's ability to generalize across diverse patient populations and imaging conditions.

\section{Problem Definition}
The main challenges in classifying thoracic diseases from chest X-ray images include dataset imbalance, multi-label classification, and image quality variations. The ChestX-ray14 dataset, for example, contains a disproportionate number of images labeled as "No Finding," which dominate the dataset and lead to biased models that are less effective in detecting rarer diseases \cite{johnson2019survey}. Additionally, the presence of multiple diseases in a single image makes multi-label classification an essential task in this context. 

Further complicating this problem is the variation in image quality due to differences in X-ray acquisition methods. In particular, images can be taken from different orientations, such as front-to-back or back-to-front, which introduces noise and reduces the consistency of the dataset. To address these challenges, we propose an approach that integrates advanced preprocessing techniques, including PCA-based image compression and class-weighted loss functions, to improve model accuracy and reduce bias.

\section{Data Collection and Preprocessing}
The ChestX-ray14 dataset consists of 112,120 X-ray images from 30,805 unique patients, labeled with one or more of 14 thoracic diseases \cite{wang2017chestxray14}. The dataset is highly imbalanced, with "No Finding" being the most frequent label. To mitigate this issue, we selected a subset of approximately 20,000 images for model training. These images were resized to 256x256 pixels to reduce memory usage and increase training efficiency.

We applied Principal Component Analysis (PCA) to compress the images, retaining 100\% of the variance with only 40 components. This compression technique significantly reduces the computational complexity while preserving the essential features of the X-ray images \cite{turk1991eigenfaces}. The resulting images were then transformed into multidimensional arrays and saved as NumPy files for use in model training.

In addition to image preprocessing, we transformed the metadata to facilitate multi-label classification. Each image in the original dataset was associated with multiple rows, each containing a different disease label. We restructured the data so that each image was represented by a single row containing all its disease labels, making it suitable for training multi-label classification models.

\begin{figure}[H]
    \centering
    \includegraphics[width=\linewidth]{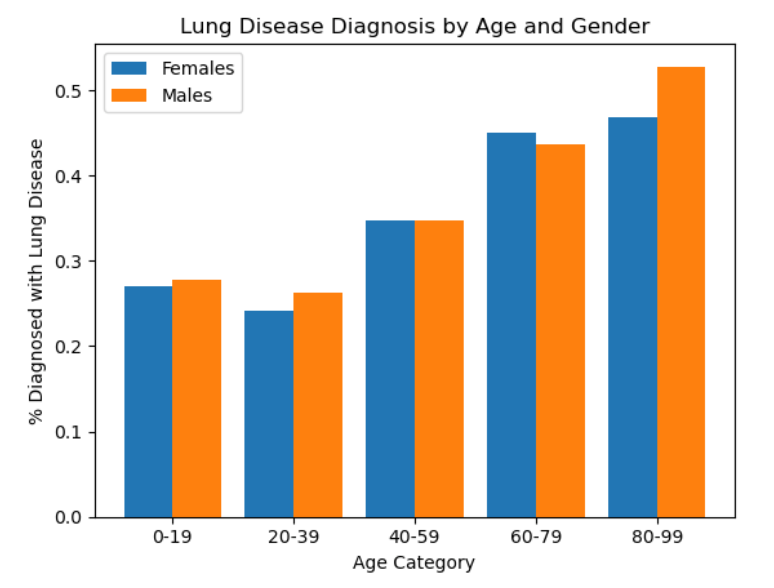}
    \caption{Patient Demographics Analysis}
    \label{fig:demographics}
\end{figure}

To determine the number of components that are required to retain enough variance, our team analyzed the compression performance of PCA. Below is the image sample that was used to visualize how the number of components affects the variance retention across the three channels.

\begin{figure}[H]
    \centering
    \includegraphics[width=\linewidth]{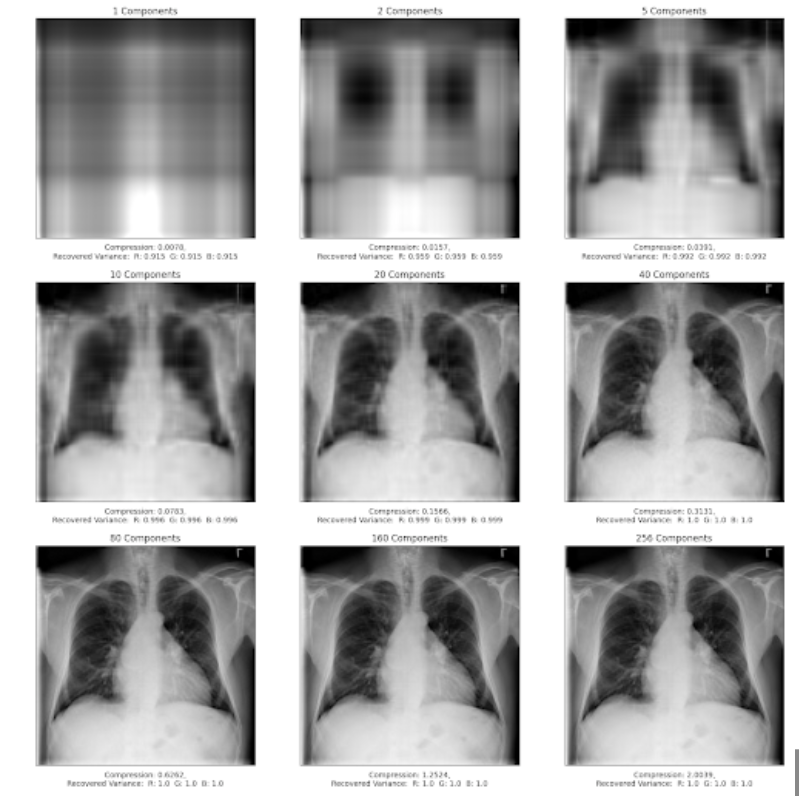}
    \caption{PCA Compression Analysis for Variance Retention}
    \label{fig:pca_compression}
\end{figure}

\section{Methodology}

\subsection{CNN Architectures}
We began our experimentation with a simple binary classification CNN model designed to classify chest X-ray images as either "No Finding" or "Disease Present." The model consisted of three convolutional layers, each with 32 filters and a kernel size of 3, followed by a dense layer with 128 nodes. The architecture was trained using the Adam optimizer with a learning rate of 0.001.

For the multi-label classification task, we extended the binary model to predict the presence of multiple diseases in a single chest X-ray image. The architecture included additional convolutional layers and fully connected layers to capture the complex relationships between the diseases. We used a sigmoid activation function in the final layer to handle the multi-label classification task.

Additionally, we explored the use of ResNet50, a deeper CNN architecture with residual connections that help mitigate the vanishing gradient problem \cite{he2016deep}. The final layers of ResNet50 were customized to output probabilities for each disease label, using a sigmoid activation function. We compared the performance of ResNet50 with the simpler CNN models to assess its ability to generalize across the multi-label task.

\subsection{Class-Weighted Loss Function}
To address the dataset imbalance, we incorporated a class-weighted loss function, which assigns higher weights to underrepresented classes during training. The weighted cross-entropy loss is defined as:

\begin{equation}
L_{\text{weighted}} = - \sum_{i=1}^{C} w_i \cdot y_i \cdot \log(p_i)
\end{equation}

where \( w_i \) is the weight assigned to class \( i \), \( y_i \) is the true label for class \( i \), and \( p_i \) is the predicted probability for class i The weights \( w_i \) are calculated as the inverse of the class frequencies:

\begin{equation}
w_i = \frac{N}{n_i}
\end{equation}

where \( N \) is the total number of samples and \( n_i \) is the number of samples in class \( i \). This loss function ensures that the model gives more importance to underrepresented diseases, improving its ability to detect rare conditions.

\subsection{Evaluation Metrics}
We evaluated model performance using accuracy, area under the curve (AUC), precision, recall, and F1-score. AUC is a more informative metric for imbalanced datasets, as it accounts for both false positives and false negatives. Precision and recall were used to assess the model's ability to detect each disease, while F1-score was used to balance the trade-off between precision and recall.

\section{Results and Discussion}

\subsection{Binary Classification Results}
The baseline binary classification model achieved 60\% accuracy and an AUC of 0.596. After optimizing the model by adding more convolutional layers and increasing the number of nodes in the dense layer, we achieved an accuracy of 66\% with an AUC of 0.708.

\begin{figure}[H]
    \centering
    \includegraphics[width=0.8\linewidth]{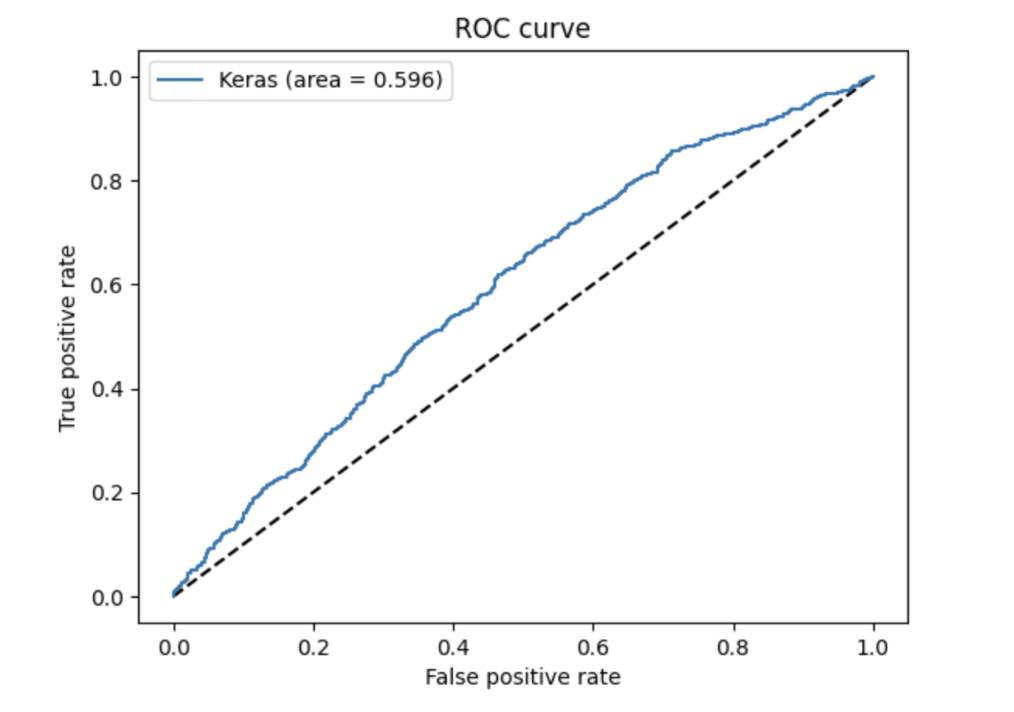} 
    \caption{ROC Curve for Baseline Binary Classification Model}
    \label{fig:roc_baseline}
\end{figure}

\begin{figure}[H]
    \centering
    \includegraphics[width=0.8\linewidth]{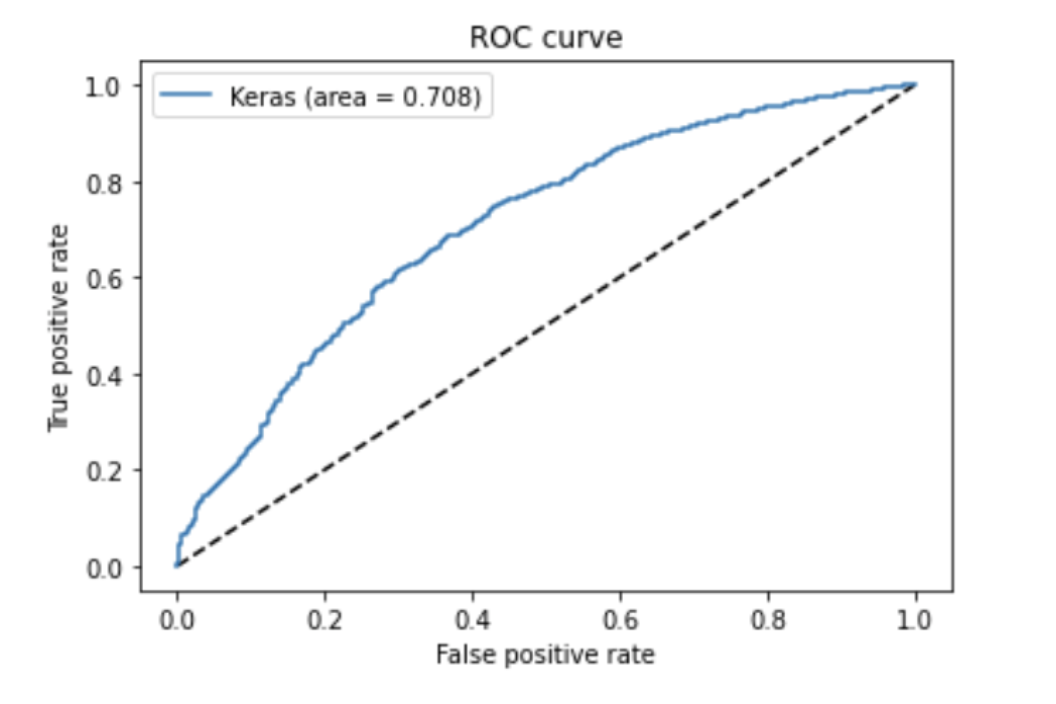} 
    \caption{ROC Curve for Optimized Binary Classification Model}
    \label{fig:roc_optimal}
\end{figure}

\subsection{Challenges in Multi-Label Classification}
The multi-label model struggled with detecting less common diseases, as expected due to the imbalance in the dataset. However, by incorporating a class-weighted loss function, we were able to slightly improve performance on underrepresented classes. Despite this, the model still had difficulties in predicting diseases that appeared visually similar, such as "Atelectasis" and "Infiltration."

\subsection{ResNet50 Model}
The ResNet50 model demonstrated promise, especially in terms of generalization to unseen data. However, due to computational constraints, we were unable to train the model fully within the time frame of this study. Preliminary results showed that ResNet50 outperformed the baseline model, particularly in terms of capturing more complex features and offering better accuracy for multi-label predictions.

\begin{figure}[H]
    \centering
    \includegraphics[width=\linewidth]{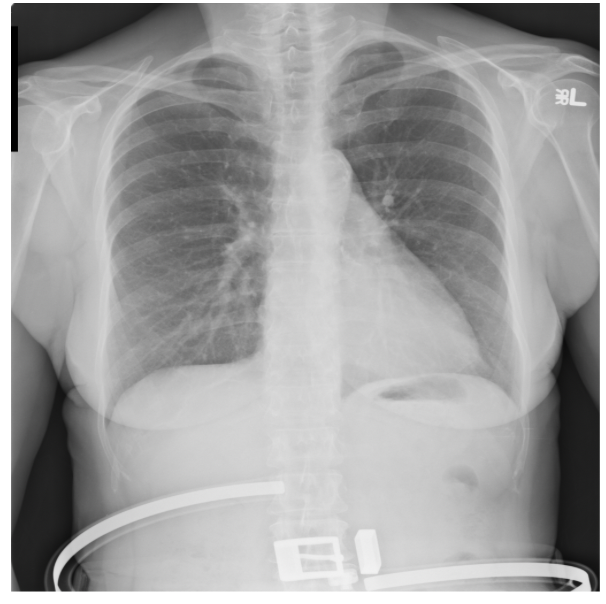}
    \caption{Example X-ray Image: Cardiomegaly}
    \label{fig:cardiomegaly}
\end{figure}

\begin{figure}[H]
    \centering
    \includegraphics[width=\linewidth]{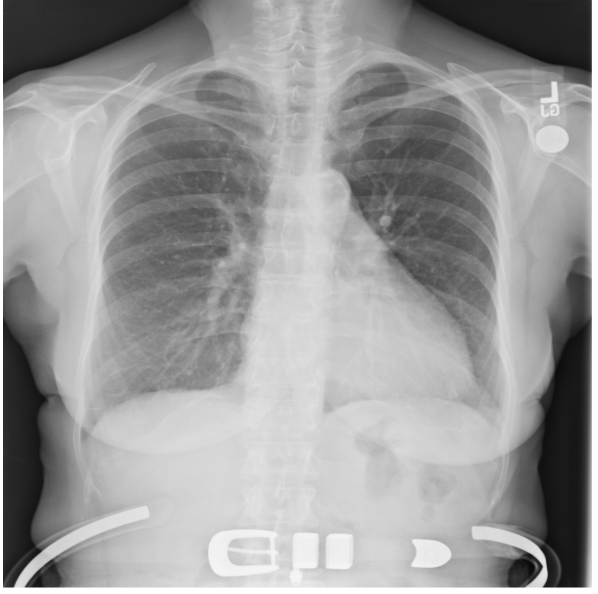}
    \caption{Example X-ray Image: No Finding}
    \label{fig:no_finding}
\end{figure}

\section{Conclusions and Future Work}
In this study, we demonstrated the feasibility of using CNNs for chest X-ray classification, focusing on both binary and multi-label classification tasks. Our optimized binary classification model achieved a respectable accuracy of 66\% and an AUC of 0.708, while our multi-label model showed the potential to classify multiple thorax diseases simultaneously, albeit with challenges due to class imbalance.

Future work will focus on improving the multi-label classification model by incorporating additional techniques such as data augmentation, synthetic data generation, and transfer learning. Furthermore, we will explore the incorporation of patient demographics and imaging conditions to improve model robustness and clinical applicability.

\end{document}